\documentclass[letterpaper]{article}
\usepackage{uai_format}
\usepackage[margin=1in]{geometry}

\usepackage{times}

\usepackage{graphicx} % more modern
\usepackage{subcaption}
\usepackage{amsmath, amsthm, amsfonts}       % blackboard math symbols
\usepackage{algorithm, algorithmic}
\usepackage{enumitem}
\usepackage{color}
\usepackage{hyperref}
\usepackage{natbib}

\newcommand{\iid}{\overset{iid}{\sim}}
\newcommand{\norm}[1]{\left\lVert#1\right\rVert}
\theoremstyle{plain}
\newtheorem*{post}{Postulate}

\let\OLDthebibliography\thebibliography
\renewcommand\thebibliography[1]{
  \OLDthebibliography{#1}
  \setlength{\parskip}{0pt}
  \setlength{\itemsep}{0pt plus 0.3ex}
}

\newcommand*\samethanks[1][\value{footnote}]{\footnotemark[#1]}

\title{Causal Inference via Kernel Deviance Measures}
\author{ {\bf Jovana Mitrovic}\thanks{now at DeepMind, UK} \\
Department of Statistics\\
University of Oxford \\
\And
{\bf Dino Sejdinovic}  \\
Department of Statistics\\
University of Oxford \\
\And
{\bf Yee Whye Teh}\samethanks \\
Department of Statistics \\
University of Oxford \\
}

\begin{document}

\maketitle

\begin{abstract}
Discovering the causal structure among a set of variables is a fundamental problem in many areas of science. In this paper, we propose Kernel Conditional Deviance for Causal Inference (KCDC) a fully nonparametric causal discovery method based on purely observational data. From a novel interpretation of the notion of asymmetry between cause and effect, we derive a corresponding asymmetry measure using the framework of reproducing kernel Hilbert spaces. Based on this, we propose three decision rules for causal discovery. We demonstrate the wide applicability of our method across a range of diverse synthetic datasets. Furthermore, we test our method on real-world time series data and the real-world benchmark dataset T\"übingen Cause-Effect Pairs where we outperform existing state-of-the-art methods.
\end{abstract}

\section{INTRODUCTION}
In many areas of science, we strive to answer questions that are fundamentally causal in nature. For example, in medicine one is often interested in the genetic drivers of diseases, while in commerce one might want to identify the motives behind customers' purchasing behaviour. Furthermore, it is of the utmost importance to thoroughly understand the underlying causal structure of the data-generating process if we are to predict, with reasonable accuracy, the consequences of interventions or answer counterfactual questions about what would have happened had we acted differently. While most machine learning methods excel at prediction tasks by successfully inferring statistical dependencies, there are still many open questions when it comes to uncovering the causal dependencies between the variables driving the underlying data-generating process. Given the growing interest in using data to guide decisions in areas where interventional and counterfactual questions abound, causal discovery methods have attracted considerable research interest \citep{hoyer2009nonlinear, zhang2009identifiability, lopez2015towards, mooij2016distinguishing}.

While causal inference is preferably performed on data coming from randomized control experiments, often this kind of data is not available due to a combination of ethical, technical and financial considerations. These real-world limitations have motivated research into inferring causal relationships from purely observational data. One group of methods \citep{spirtes2000causation, sun2007kernel} attempts to recover the causal structure by analyzing conditional independencies present in the data, but does not provide a definitive answer for the underlying causal structure and is not robust to the choice of conditional independence testing methodology. Another group of methods \citep{hoyer2009nonlinear, zhang2009identifiability, mooij2009regression} postulates that there is some inherent asymmetry between cause and effect and proposes different asymmetry measures that form the basis for causal discovery. While these methods provide a definitive answer to the question of causal structure, they typically assume a particular functional form for the interaction between the variables and a particular noise structure which limits their applicability. We aim our contribution to be a step towards a method that can deal with highly complex data-generating processes, provides a definitive answer for the causal structure relying only observational data and whose inference can easily be extended without the need to develop novel, specifically tailored algorithms for each new model class.

\begin{figure}[!h]
\begin{subfigure}{.24\textwidth}
  \centering
  \includegraphics[width=.9\linewidth]{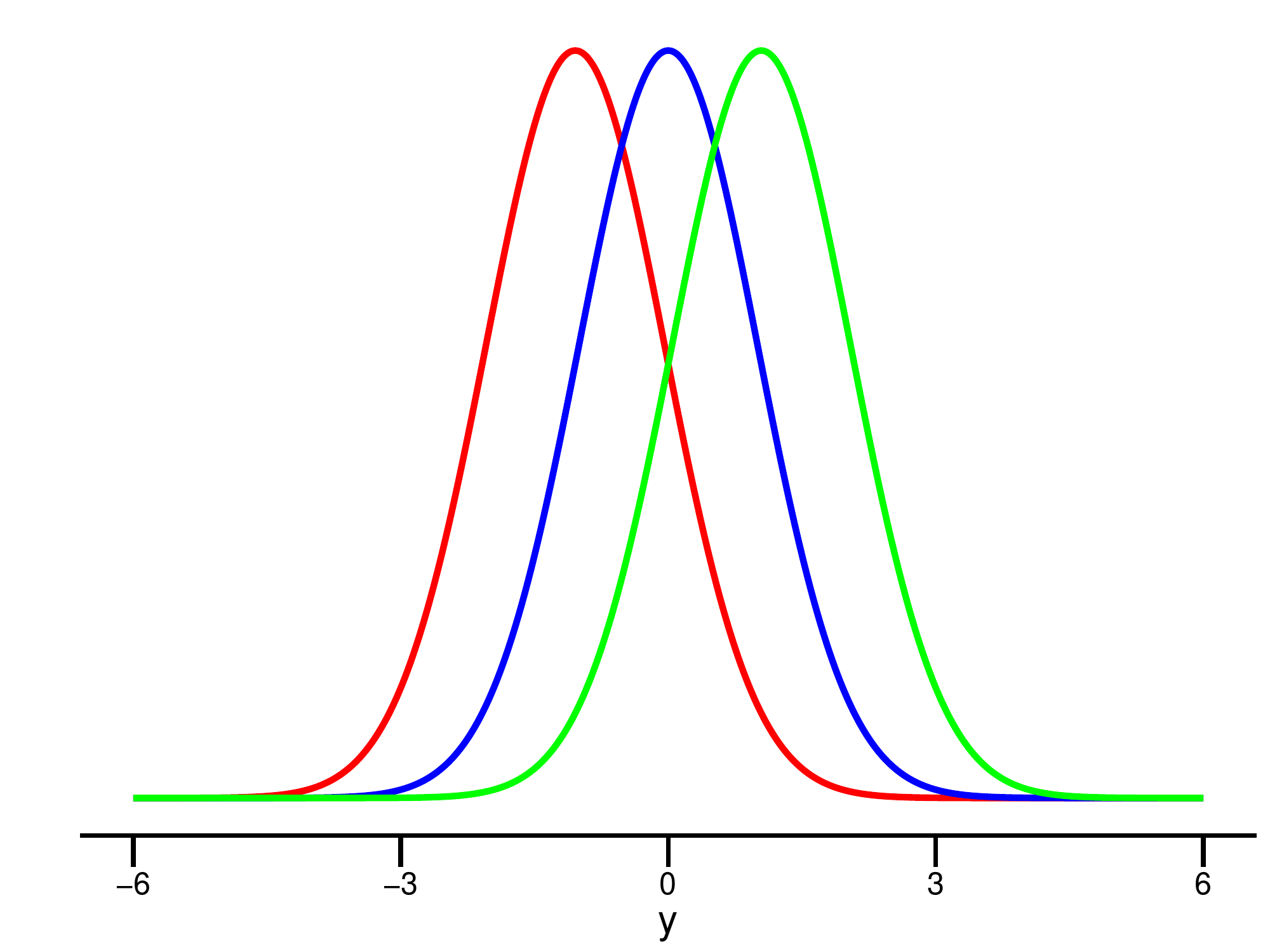}
\end{subfigure}%
\begin{subfigure}{.24\textwidth}
  \centering
  \includegraphics[width=.9\linewidth]{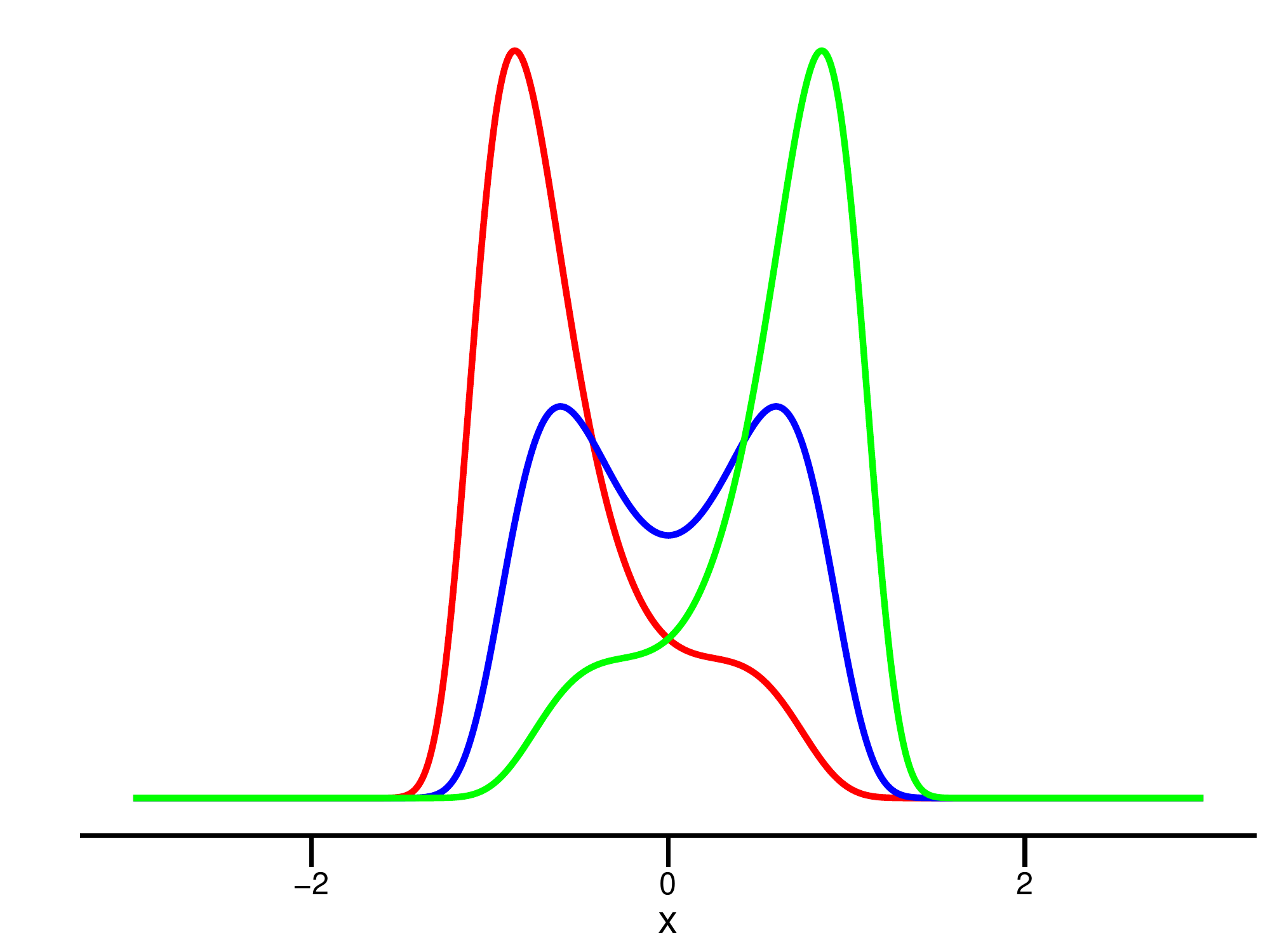}
\end{subfigure}
\caption{Causal And Anticausal Direction, Respectively.} \label{example}
\end{figure}

In this work, we develop a fully nonparametric causal inference method to automatically discover causal relationships from purely observational data. In particular, our proposed method does not require any \emph{a priori} assumptions on the functional form of the interaction between the variables or the noise structure. Furthermore, we propose a novel interpretation of the notion of asymmetry between cause and effect \citep{daniusis2012inferring}. Before we introduce our proposed interpretation, we motivate it with the following example. Let $y = x^{3} + x + \epsilon$ with $\epsilon\sim\mathcal{N}(0,1)$ where we consider the correct causal direction to be $x\rightarrow y$. Figure \ref{example} visualizes the conditional distributions $p(y\vert x)$ and $p(x\vert y)$ for different values of $x$ and $y$, respectively.

Note that the conditional distributions in the anticausal direction exhibit a larger structural variability across different values of the conditioning variable than the conditional distributions in the causal direction. It is important to note here that structural variability does not only refer to variability in the scale and location parameters, but should be understood more broadly as variability in the ``parametric'' form, e.g. differences in the number of modes and in higher order moments. If one thinks of conditional distributions as programs generating $y$ from $x$ and vice versa, we see that in the causal direction the structure of the program remains unchanged although different input arguments are provided. In the anticausal direction, we see that the program requires structural modification across different values of the input in order to account for the differing behaviour of the conditional densities.

Motivated by the above observation, we popose a novel interpretation of the notion of asymmetry between cause and effect in terms of the shortest description length, i.e. Kolmogorov complexity \citep{grunwald2008algorithmic}, of the data-generating process. Whereas previous work \citep{lemeire2006causal, janzing2010causal,daniusis2012inferring, Budhathoki17} quantifies the asymmetry in terms of the Kolmogorov complexity of the factorization of the joint distribution, we propose to interpret the asymmetry based on the Kolmogorov complexity of the conditional distribution.
Specifically, we propose that this asymmetry is realized by \emph{the Kolmogorov complexity of the mechanism in the causal direction being independent of the input value of the cause}. On the other hand, in the anticausal direction, there will be a dependence between the shortest description length of the mechanism and the particular input value of the effect. This (in)dependence can be measured by looking at the variability of Kolmogorov complexities of the mechanism for particular of the input. Unfortunately, as computing the Kolmogorov complexity is an intractable problem, we resort to conditional distributions as approximations of the corresponding programs. Thus, we can infer the causal direction by comparing the description length variability of conditional distributions across different values of the conditioning variable with the causal direction being the less variable. For measuring this variability, we use the framework of reproducing kernel Hilbert spaces (RKHS). This allows us to represent conditional distributions in a compact, yet expressive way and efficiently capture their many nuanced aspects thus enabling more accurate causal inference. In particular, by way of the kernel trick, we can efficiently compute the variability of infinite-dimensional objects using finite-dimensional quantities that can be easily estimated from data. Using the RKHS framework makes our method readily applicable also in situations when trying to infer the causal direction between pairs of random variables taking values in structured or non-Euclidean domains on which a kernel can be defined. Next, we propose three decision rules for causal inference based on the description length variability of sets of conditional distributions.

The main contributions of this paper are:
\begin{itemize}
 \item an interpretation of the notion of asymmetry between cause and effect in terms of the independence of the description length of the mechanism on the value of the cause,
 \item an approximation to the intractable description length in terms of conditional distributions,
 \item a flexible asymmetry measure based on RKHS embeddings of conditional distributions,
 \item a fully nonparametric method for causal inference that does not impose \emph{a priori} any assumptions on the functional relationship between the variables or the noise structure.
\end{itemize}

The rest of the paper is organized as follows. In Section 2, we review related work, while in Section 3 we introduce and discuss our causal inference methodology. In Section 4, we present experimental results on synthetic and real-world datasets. We discuss extensions to the case of more than two variables in Section 5. In Section 6, we discuss future research directions and conclude.

\section{RELATED WORK}
Most approaches to causal inference from purely observational data can be grouped into three
categories. The first category of approaches, so-called constraint-based methods, assume that the true causal structure can be represented with a directed acyclic graph (DAG) and then try to infer the true causal DAG $G$ by analyzing conditional independencies present in the observational data distribution $P$. Under some technical assumptions on the relationship between $G$ and $P$ \citep{judea2000causality},  these methods can determine $G$ up to its Markov equivalence class\footnote{All DAGs that encode the same set of conditional independence relations constitute a Markov equivalence class.} which usually contains DAGs that can be structurally very diverse and still have many unoriented edges thus not allowing for a definitive answer to the question of causal structure. An example of this approach is the PC algorithm \citep{spirtes2000causation} which builds a graph skeleton by successively removing unnecesary connections between the variables and then orienting the remaining edges if possible. Other examples of this approach rely on kernel-based conditional independence criteria, e.g.  \citep{sun2007kernel, zhang2011kernel}. Although mathematically well-founded, the performance of these methods is highly dependent on the utilized conditional independence methodology, whose performance usually depends on the amount of available data. Furthermore, these methods are not robust as small errors in building the graph skeleton (e.g. a missing independence relation) can lead to significant errors in the inferred Markov equivalence class. As conditional independence tests require at least three variables, they are not applicable in the two variable case.

A second class of models, so-called score-based methods, searches the space of all DAGs of a certain size by scoring their fit to the observed data using a predefined score function. An example of this approach is Greedy Equivalent Search \citep{chickering2002optimal} which combines greedy search with the Bayesian information criterion. As the search space grows super-exponentially with the number of variables, these methods quickly become computationally intractable. An answer to this shortcoming are hybrid methods which use constraint-based approaches to decrease the search space that can then be effectively explored with score-based methods, e.g. \citep{tsamardinos2006max}. DAGs have also been represented using generative neural networks and scored according to how well the generated data matches the observed data, e.g. \citep{goudet2017causal}. A major shortcoming of this hybrid methodology is that there exists no principled way of choosing problem-specific combinations of scoring functions and search strategies which is a significant problem as different search strategies in combination with different scoring rules can potentially lead to very different results.

The third category of methods assumes that there exists some inherent asymmetry between cause and effect. One line of research, often refered to as functional causal models or structural equation models, assumes a particular functional form for the causal interactions between the variables and a particular noise structure. In these models, each variable is a deterministic function of its causes and some independent noise, with all noise variables assumed to be jointly independent. Examples of this methodology assume linearity and additive non-Gaussian noise \citep{shimizu2006linear}, nonlinear additive noise \citep{hoyer2009nonlinear, mooij2009regression} and invertible interactions between the covariates and the noise \citep{zhang2009identifiability}. In order to perform causal discovery in these models, the special structural assumptions placed on the interaction between the covariates and on the noise are of crucial importance, thus limiting their applicability. A second strand of research interprets the asymmetry between cause and effect through an information-theoretic lens by examining the complexity of the factorization of the joint distribution \citep{lemeire2006causal}. \citep{janzing2010causal} argue that if $X$ causes $Y$, then the factorization in the causal direction, i.e. $p(X,Y)=p(Y\vert X)p(X)$, should have a shorter description in terms of the Kolmogorov complexity
than the factorization in the anticausal direction, i.e. $p(X,Y)=p(X \vert Y)p(Y)$. In \citep{daniusis2012inferring}, instead of computing the intractable Kolmogorov complexity, the correlation between the input and the conditional distribution is measured, whereas \citep{Budhathoki17} use the minimum description length principle. The approach of \citep{sun2007distinguishing} measures the complexity of conditional distributions by RKHS seminorms computed on the logarithms of their densities.

Lastly, causal discovery has also been framed as a learning problem. Examples of this approach are \citep{lopez2015towards, fonollosa2016conditional}. RCC \citep{lopez2015towards} constructs feature representations of the data based on RKHS embeddings of the joint and marginal distributions and uses these within a random forest classifier. In \citep{fonollosa2016conditional}, the feature representation of the data includes quantities describing the joint, marginal and conditional distributions. In particular, the conditional distributions are represented with conditional entropy, mutual information and a quantification of their variability in terms of the spread of the entropy, variance and skewness for different values of the conditioning variable.

On the other hand, we propose a causal inference methodology based on a novel interpretation of the asymmetry between cause and effect and derive three decision rules with one of these decision rules based on classifying feature representation of the data. In particular, we consider feature representations based only on conditional distributions which we argue to be more discriminative for inferring the causal direction.

\section{KERNEL CONDITIONAL DEVIANCE FOR CAUSAL INFERENCE}
We first briefly review some basics of RKHS theory that constitute the building blocks of our approach. For a detailed discussion, see \citep{scholkopf2001learning}.

\subsection{BACKGROUND}
Let $(\mathcal{X}, \mathcal{B_{X}})$ and $(\mathcal{Y},\mathcal{B_{Y}})$ be measurable spaces with $\mathcal{B_{X}}$ and $\mathcal{B_{Y}}$ the associated Borel $\sigma$-algebras. Denote by $(\mathcal{H_{X}}, k)$ and $(\mathcal{H_{Y}}, l)$ the RKHSs of functions defined on $\mathcal{X}$ and $\mathcal{Y}$, respectively, and their corresponding kernels. Given a probability distribution $p$ on $\mathcal{X}$, the \emph{mean embedding} $\mu_{p}$\footnote{$\mu_{p}$ and $\mu_{X}$ will be used interchangeably if it does not lead to confusion.}   \citep{scholkopf2001learning} is a representation of $p$ in $\mathcal{H_{X}}$ given by
\begin{equation*}
\mu_{p} = \mathbb{E}_{p}[ k(\cdot, X) ]
\end{equation*}
with $X\sim p$. This can be unbiasedly estimated by
\begin{equation*}
\hat{\mu}_{p} = \frac{1}{n} \sum_{i=1}^{n} k(\cdot, x_{i})
\end{equation*}
with $\{x_{i}\}_{i=1}^{n}\iid p$. Furthermore, if $k$ is a characteristic kernel \citep{scholkopf2001learning}, then this representation yields a metric on probability measures, i.e. $\norm{ \mu_{p} - \mu_{q} }_{\mathcal{H}_{k}}= 0 \Leftrightarrow p = q$. The radial basis function (RBF) kernel with bandwidth $\sigma$ given by
\begin{equation*}
    k(x, x^{\prime}) = \exp\left( -\frac{\norm{ x-x^{\prime} }^{2}}{2\sigma^{2}} \right)
\end{equation*}
is an example of a characteristic kernel. A conditional distribution $p(X\vert Y=y)$ can be encoded using the \emph{conditional mean embedding} $\mu_{X\vert Y=y}$ \citep{scholkopf2001learning} which is an element of $\mathcal{H_{X}}$ that satisfies
\begin{equation*}
    \mathbb{E}[h(X)\vert Y=y] = \langle h, \mu_{X\vert Y=y} \rangle_{\mathcal{H_{X}}} \quad \forall h\in\mathcal{H_{X}}.
\end{equation*}
Using the equivalence between conditional mean embeddings and vector-valued regressors  \citep{lever2012conditional}, we can estimate $\mu_{X\vert Y=y}$ from a sample $\{(x_{i}, y_{i})\}_{i=1}^{n}\iid p(x,y)$ as
\begin{align}
    \begin{split}\label{cond_mean}
    &\hat{\mu}_{X\vert Y=y} =  \sum_{i=1}^{n} \alpha_{i}(y) k(\cdot, x_{i}) \\
    & \text{with} \quad \alpha(y) = ({\bf L} + n\lambda {\bf I})^{-1}{\bf l}_{y},
    \end{split}
\end{align}
with ${\bf L} = [l(y_{i}, y_{j})]_{i,j=1}^{n}$, ${\bf l}_{y} = [l(y_{1}, y), \dots, l(y_{n}, y)]^{T}$, $\alpha(\cdot) = [\alpha_{1}(\cdot), \dots, \alpha_{n}(\cdot)]^{T}$, regularization parameter $\lambda$ and identity matrix ${\bf I}$.

\subsection{METHOD}
For simplicity, we restrict our attention to the two variable problem of causal discovery, i.e. distinguishing between cause and effect. Possible extensions to the multivariable setting are discussed in Section 5. Following the usual approach in the literature, we derive our method under the assumption of causal sufficiency of the data. In particular, we ignore the potential existence of confounders, i.e. all causal conclusions should be understood with respect to the set of observed variables. Nevertheless, in Section 4, we see that our method performs well also in settings where the noise has positive mean which can be interpreted as accounting for potential confounders.

Given observations $\{(x_{i}, y_{i})\}_{i=1}^{n}$ of a pair of random variables $(X,Y)$, our goal is to infer the causal direction, i.e.\ decide whether $X$ causes $Y$ (i.e. $X\rightarrow Y$)  or $Y$ causes $X$ (i.e. $Y \rightarrow X$). To this end, we develop a fully nonparametric causal discovery method that relies only on observational data. In particular, our method does not \emph{a priori} postulate a particular functional model for the interactions between the variables or a particular noise structure. Our approach, Kernel Conditional Deviance for Causal Inference (KCDC), is based on the assumption that there exists an asymmetry between cause and effect that is inherent in the data-generating process.
While there are many interpretations of how this asymmetry might be realized, two of the more prominent ideas phrase it in terms of the independence of cause and mechanism \citep{daniusis2012inferring} and in terms of the complexity of the factorization of the joint distribution \citep{lemeire2006causal, janzing2010causal}.

Motivated by these two ideas, we propose a novel interpretation of the notion of asymmetry between cause and effect. First, we take an information-theoretic approach to reasoning about the complexity of distributions similar to \citep{lemeire2006causal, janzing2010causal}. In particular, we reason about it in terms of algorithmic complexity, i.e. Kolmogorov complexity \citep{grunwald2008algorithmic} which is the description length of the shortest program that implements the sampling process of the distribution. For a distribution $p(Y)$, the Kolmogorov complexity is
\begin{equation*}
 K(p(Y)) = \min_{s} \{\vert s\vert : \vert U(s, y, q) - p(y) \vert \leq q \;\; \forall y\}
\end{equation*}
with $q$ a precision parameter, $U(\cdot)$ extracting the output of applying program $s$ onto a realization of the random variable $Y$ denoted by $y$. Analogously, for a conditional distribution $p(Y\vert X)$, the Kolmogorov complexity is
\begin{align*}
 K(p(Y\vert X)) = & \\
      \min_{s} \{\vert s\vert &: \vert U(s, y, x, q) - p(y\vert x) \vert \leq q \;\; \forall x, y\}.
\end{align*}
Assuming $X\rightarrow Y$, the asymmetry notion specified in terms of factorization complexity can be expressed as
\begin{equation*}
 K(p(X)) + K(p(Y\vert X)) \leq K(p(Y)) + K(p(X\vert Y))
\end{equation*}
which holds up to an additive constant \citep{stegle2010probabilistic}. Further, the independence of cause and mechanism can be interpreted as algorithmic independence \citep{janzing2010causal}, i.e. knowing the distribution of the cause $p(X)$ does not enable a shorter description of the mechanism $p(Y\vert X)$.

Based on this, we argue that not only knowing the distribution of the cause does not enable a shorter description of the mechanism, but also knowing any particular value of the cause does not provide any information that can be used to construct a shorter description of the mechanism. To formalize this, we introduce the notation
\begin{align*}
K(p(Y\vert X=x)) = & \\ \min_{s} \{\vert s\vert :
\vert U(s, & y, x, q) -  p(y\vert X=x) \vert \leq q  \;\; \forall y\}
\end{align*}
to be the Kolmogorov complexity of the conditional distribution $p(Y\vert X)$ when the conditioning variable takes on the value $X=x$. From our argument above, we see that in the causal direction the Kolmogorov complexity of $p(Y\vert X=x)$ is independent of the particular value $x$ of the cause $X$, i.e.
\begin{equation*}
    K(p(Y \vert X=x_{i})) =  K(p(Y \vert X=x_{j})) \quad \forall i, j.
\end{equation*}
On the other hand, this will not hold in the anticausal direction as the input and mechanism are not algorithmically independent in that direction, i.e.
\begin{equation*}
    K(p(X \vert Y=y_{i})) \neq  K(p(X \vert Y=y_{j})) \quad \forall i\neq j.
\end{equation*}

This motivates our interpretation of the notion of asymmetry between cause and effect
which is summarized as follows.

\begin{post}
(Minimal description length independence) \\ If $X\rightarrow Y$, the minimal description length of the mechanism mapping $X$ to $Y$ is independent of the value of $X$, whereas the minimal description length of the mechanism mapping $Y$ to $X$ is dependent on the value of $Y$.
\end{post}

Building on this, we can infer the causal direction by comparing how much the description length of the minimal description length program implementing the mechanism varies across different values of its input arguments. In particular, in the causal direction, we expect to see less variability than in the anticausal direction. As computing the Kolmogorov complexity is an intractable problem, we use th norm of RKHS embeddings of the corresponding conditional distributions as a proxy for it. Thus, we recast causal inference in terms of comparing the variability in RKHS norm of embeddings of sets of conditional distributions indexed by values of the conditioning variable. In order to perform causal inference, we use the framework of reproducing kernel Hilbert spaces. This allows us to construct highly expressive, yet compact approximations of the potentially highly-complex programs and circumvent the challenges of density estimation when trying to represent conditional distributions. Furthermore, using the RKHS framework allows us to efficiently capture the many nuanced aspects of distributions thus enabling more accurate causal inference. For example, using non-linear kernels allows us to capture more comprehensive distributional properties including higher order moments. Furthermore, using the RKHS framework makes our method readily applicable also in situations when trying to infer the causal direction between two random vectors (treated as single variables) or pairs of other types of random variables taking values in structured or non-Euclidean domains on which a kernel can be defined. Examples of such types of data include discrete data, genetic data, phylogenetic trees, strings, graphs and other structured data \citep{gartner2002kernels}.

We represent conditional distributions in the RKHS using conditional mean embeddings \citep{scholkopf2001learning}. In particular, given observations $\{(x_{i}, y_{i})\}_{i=1}^{n}$ of a pair of random variables $(X,Y)$, we construct the embeddings of the two sets of conditional distributions, $\{p(X\vert Y=y_{i})\}_{i=1}^{n}$ and
$\{p(Y\vert X=x_{i})\}_{i=1}^{n}$ using (\ref{cond_mean}). Furthermore, if we choose a characteristic kernel \citep{scholkopf2001learning}, the conditional mean embeddings of two distinct distributions will not overlap. For example, we can choose the RBF kernel which is characteristic and embeds the distributions into the Hilbert space of infinitely differentiable functions. Next, we compute the variability in RKHS norm of a set of conditional mean embeddings as the deviance of the RKHS norms of that set. Thus, using the KCDC measure $S_{X\rightarrow Y}$ with
\begin{align}
  \begin{split}\label{kcdc1}
  S_{X\rightarrow Y} = \frac{1}{n}\sum_{i=1}^{n} \Bigg(&\norm{\mu_{Y\vert X=x_{i}}}_{\mathcal{H_{Y}}} \\ & \quad
                     - \frac{1}{n}\sum_{j=1}^{n} \norm{\mu_{Y\vert X=x_{j}}}_{\mathcal{H_{Y}}} \Bigg)^{2},
  \end{split}
\end{align}
we compute the deviance in RKHS norm of the set $\{p(Y\vert X = x_{i})\}_{i=1}^{n}$ .
Analogously, for $\{p(X\vert Y=y_{i})\}_{i=1}^{n}$, the KCDC measure can be calculated as
\begin{align}
 \begin{split}\label{kcdc2}
  S_{Y\rightarrow X} = \frac{1}{n} \sum_{i=1}^{n}  \Bigg(&\norm{\mu_{X\vert Y=y_{i}}}_{\mathcal{H_{X}}} \\ & \quad
                      - \frac{1}{n} \sum_{j=1}^{n} \norm{\mu_{X\vert Y=y_{j}}}_{\mathcal{H_{X}}} \Bigg)^{2}
 \end{split}
\end{align}

Based on our proposed interpretation of the notion of asymmetry between cause and effect, we can determine the causal direction between $X$ and $Y$. For this purpose, we propose three different decision rules. First, we can determine the causal direction by directly comparing the KCDC measures for the two directions, i.e.
\begin{equation*}
 \medmuskip=0mu
 \thinmuskip=0mu
 \thickmuskip=0mu
D_{1}(X,Y) =
\begin{cases}
    X\rightarrow Y, & \text{if} \quad S_{X\rightarrow Y} < S_{Y\rightarrow X}   \\
    Y\rightarrow X, & \text{if} \quad S_{X\rightarrow Y} > S_{Y\rightarrow X}  \\
\end{cases}
\end{equation*}
but leave the causal direction undetermined if $\frac{\vert S_{X\rightarrow Y} - S_{Y\rightarrow X}\vert}{\min(S_{X\rightarrow Y}, S_{Y\rightarrow X})} < \delta$ with $\delta$ some fixed decision threshold. The case of undetermined direction accounts for situations where the KCDC measures are too close in value to determine the causal direction. This situation might come about due to numerical errors or non-identifiability. Furthermore, we can also derive a confidence measure $\mathcal{T^{KCDC}}$ for the inferred causal direction as
\begin{equation*}
 \mathcal{T^{KCDC}} = \frac{\vert S_{X\rightarrow Y} - S_{Y\rightarrow X}\vert}{\min(S_{X\rightarrow Y}, S_{Y\rightarrow X})}
\end{equation*}

Second, we can determine the causal direction based on majority voting of an ensemble constructed using different model hyperparameters, i.e.
\begin{equation*}
D_{2}(X,Y) = \texttt{Majority}(\{D_{1}^{H_{j}}(X, Y)\}_{j})
\end{equation*}
where the dependence on the model hyperparameters $H_{j}$ has been made explicit. Third, the KCDC measures can also be used for constructing feature representations of the data which can then be used within a classification method. In particular, we can infer the causal relationship between $X$ and $Y$ using
\begin{equation*}
D_{3}(X,Y) = \texttt{Classifier}(\{S_{X\rightarrow Y}^{H_{j}}, S_{Y\rightarrow X}^{H_{j}}\}_j)
\end{equation*}
where \texttt{Classifier} is a classification algorithm that classifies $X\rightarrow Y$ against $Y\rightarrow X$. For training the classifier, we generate synthetic data, e.g. as in \citep{lopez2015towards}. The following algorithms summarize our causal inference methodology.

\begin{algorithm}[H]
   \caption{KCDC Algorithm}
   \label{kcdc}
\begin{algorithmic}
   \STATE {\bfseries Input:} Realizations $\{(x_{i}, y_{i})\}_{i=1}^{n}$ of $(X,Y)$
   \STATE {\bfseries Output:} Causal direction ($X\rightarrow Y$ vs. $Y\rightarrow X$)
   \STATE
   \STATE Determine causal direction using one of the following
   \begin{enumerate}[label=(\Alph*)]
    \item Compute $S_{X\rightarrow Y}$ and $S_{Y\rightarrow X}$ using Algorithm \ref{measure} \\
    Perform direct comparison with decision rule $D_{1}$
    \item Compute $S_{X\rightarrow Y}$ and $S_{Y\rightarrow X}$ using Algorithm \ref{measure} for different model hyperparameters $\{H_{j}\}_{j}$ \\
    Perform majority voting with decision rule $D_{2}$
    \item Compute $S_{X\rightarrow Y}$ and $S_{Y\rightarrow X}$ using Algorithm \ref{measure} for different model hyperparameters $\{H_{j}\}_{j}$ \\
    Build data representation with $\{S_{X\rightarrow Y}^{H_{j}}, S_{Y\rightarrow X}^{H_{j}}\}_j$ \\
    Train \texttt{Classifier} using synthetic data \\
    Use decision rule $D_{3}$
   \end{enumerate}
\end{algorithmic}
\end{algorithm}

\begin{algorithm}[H]
   \caption{Compute KCDC Measures}
   \label{measure}
\begin{algorithmic}
   \STATE {\bfseries Input:} Realizations $\{(x_{i}, y_{i})\}_{i=1}^{n}$ of $(X,Y)$, \\ \hspace{1cm} kernel hyperparameters $H_{j}$
   \STATE {\bfseries Output:} KCDC measures $S_{X\rightarrow Y}$ and $S_{Y\rightarrow X}$
   \STATE
   \FOR{$i=1, \dots, n$}
   \STATE Embed $\{p(Y\vert X=x_{i})\}_{i=1}^{n}$ and $\{p(X\vert Y=y_{i})\}_{i=1}^{n}$ with (\ref{cond_mean}) using kernel hyperparameters $H_{j}$
   \ENDFOR
   \vspace{0.1cm}
   \STATE Compute $S_{X\rightarrow Y}$ and $S_{Y\rightarrow X}$ using (\ref{kcdc1}) and (\ref{kcdc2})
\end{algorithmic}
\end{algorithm}

{\bf Identifiability.} For methods that assume a functional model and determine the causal direction based on the independence between covariates and noise, \citep{zhang2009identifiability} show that the assumed functional class needs to be constrained in order to ensure the identifiability of the model. Although KCDC is not based on this approach, it still fulfills the above requirement as the kernel hyperparameters used for computing the KCDC measures are the same in both directions. Given our approach to causal inference, the causal direction will not be identifiable in situations where the description length of conditional distributions in both the causal and anticausal direction does not vary with the value of the cause and effect, respectively. This happens when in both directions the functional form of the mechanism can be described by one family of distributions for all its input arguments. One example of this is linear Gaussian dependence which is non-identifiable for most other causal discovery methods too. Another example is the case of independent variables which is usually not considered in the literature, but can be easily mitigated with an independence test. Note that using characteristic kernels eliminates any potential non-identifiability that might arise as a consequence of the non-injectivity of the embedding process.

\section{EXPERIMENTAL RESULTS}
\subsection{SYNTHETIC DATA}
In order to showcase the wide applicability of our proposed approach, we test it extensively on several synthetic datasets spanning a wide range of functional dependencies between cause and effect and different interaction patterns with different kinds of noise. We compare our approach to LiNGAM \citep{shimizu2006linear}, IGCI \citep{daniusis2012inferring}, ANM \citep{mooij2016distinguishing} with Gaussian Process regression and HSIC test \citep{scholkopf2001learning} on the residual and the post-nonlinear model (PNL) \citep{zhang2009identifiability} with HSIC test. In all of the below experiments, we sample $100$ datasets of $100$ observations each with $x\sim\mathcal{N}(0,1)$ and test three different noise regimes showcasing the robustness of our method with respect to different types of noise across different functional dependencies. In particular, the noise $\epsilon$ is either drawn from a standard normal $\mathcal{N}(0,1)$, a uniform $\mathcal{U}(0,1)$ or an exponential $\text{Exp}(1)$ distribution. Note that the exponential noise has positive mean which can be interpreted as accounting for confounders. In all experiments, we apply the decision rule based on direct comparison for KCDC. We tested across different combinations of characteristic kernels (RBF, log and rational quadratic kernels) which yielded fairly consistent performance.
In the following tables, we report the results when using the log kernel $l(x, x^{\prime}) = -\log(\norm{x - x^{\prime}}^{2} + 1)$ on the input and the rational quadratic kernel $k(x, x^{\prime}) = 1 - \frac{\norm{x - x^{\prime}}^{2}}{\norm{x - x^{\prime}}^{2} + 1}$ on the response.

{\bf Additive Noise.} As a first proof of concept, we examine the performance of our method on additive noise models as such models are the basis of many causal inference methods, e.g. \citep{hoyer2009nonlinear}. We test our approach on
(A) $y = x^{3} + x + \epsilon$,
(B) $y = \log(x + 10) + x^{6} + \epsilon$, \\
(C) $y = \sin(10x) + \exp(3x) + \epsilon$. From the table below, we see that LiNGAM performs does not perform well which is to be expected given its assumption of linear dependence. ANM performs very well across all functional and noise settings due to its assumption of additive noise. PNL does not perform well in any settting which is probably due to overfitting. IGCI peforms well for (C) and under exponential noise, while KCDC correctly classifies every dataset in every setting.
\begin{table}[h]
\caption{Additive Noise: Classification Accuracies Over $100$ Datasets.}
\label{additive_acc}
\begin{center}
\begin{tabular}{l r r r}
\hline
(A)  & Gaussian  & Uniform  & Exponential \\
\hline
LiNGAM    & 26\% & 87\% & 28\%  \\
ANM       & {\bf 100\%} & {\bf 100\%} & {\bf 100\%}  \\
PNL       & 53\% & 14\% & 47\% \\
IGCI      & 52\% & 52\% & 94\%\\
KCDC      & {\bf 100\%} & {\bf 100\%} & {\bf 100\%}  \\
\hline
(B)  & Gaussian  & Uniform  & Exponential \\
\hline
LiNGAM  & 4\%  & 40\%  & 4\%  \\
ANM     & 94\% & 97\% & 79\% \\
PNL     & 54\% & 33\% & 46\% \\
IGCI    & 54\% & 68\% & 96\%  \\
KCDC    & {\bf 100\%} & {\bf 100\%} & {\bf 100\%}  \\
\hline
(C)  & Gaussian  & Uniform  & Exponential \\
\hline
LiNGAM  & 25\% & 32\% & 18\%  \\
ANM     & 98\% & {\bf 100\%} & 97\% \\
PNL     & 39\% & 27\% & 36\% \\
IGCI    & 98\% & {\bf 100\%} & 99\% \\
KCDC    & {\bf 100\%} & {\bf 100\%} & {\bf 100\%}  \\
\hline
\end{tabular}
\end{center}
\end{table}

{\bf Multiplicative Noise.} Next, we look at datasets where the noise interacts multiplicatively with the covariates. To test our method in this setting, we generate data according to the following functional dependencies \\
(A) $y = (x^{3} + x)\exp(\epsilon)$, \\
(B) $y = (\sin(10x) + \exp(3x))\exp(\epsilon)$, \\
(C) $y = (\log(x + 10) + x^{6})\exp(\epsilon)$. \\
In the table below, we see that ANM and LiNGAM do not perform well which is to be expected given their assumption of additive noise. PNL has somewhat better performance, but does not surpass chance level in half the settings. On the other hand, IGCI peforms very well across all settings, while KCDC correctly classifies every dataset in every setting.

\begin{table}[!htbp]
\caption{Multiplicative Noise: Classification Accuracies Over $100$ Datasets.}
\label{mult_acc}
\begin{center}
\begin{tabular}{l r r r}
\hline
(A)  & Gaussian  & Uniform  & Exponential \\
\hline
LiNGAM    &  20\% & 30\% & 5\%   \\
ANM       &  0\% & 0\% & 1\%  \\
PNL       & 52\% & 24\% & 30\%  \\
IGCI      &  {\bf 100\%} & 89\%  & {\bf 100\%}   \\
KCDC      & {\bf 100\%} & {\bf 100\%} & {\bf 100\%}  \\
\hline
(B)  & Gaussian  & Uniform  & Exponential \\
\hline
LiNGAM  & 10\% & 22\% & 4\%  \\
ANM     & 8\% & 30\% & 12\%  \\
PNL     &  49\% & 58\% &  32\%  \\
IGCI    &  {\bf 100\%} & 89\%  & {\bf 100\%}   \\
KCDC    & {\bf 100\%} & {\bf 100\%} & {\bf 100\%}  \\
\hline
(C)  & Gaussian  & Uniform  & Exponential \\
\hline
LiNGAM  & 0\% & 3\% & 0\%   \\
ANM     &  5\% & 1\% & 0\%  \\
PNL     &  55\% & 41\%  & 30\%  \\
IGCI    &  {\bf 100\%} & 99\% & {\bf 100\%}  \\
KCDC    & {\bf 100\%} & {\bf 100\%} & {\bf 100\%}  \\
\hline
\end{tabular}
\end{center}
\end{table}

{\bf More complex noise.} Further, we examine exponential and periodic interactions of the covariates with the noise. In particular, we generate synthetic data according to \\
(A) $y = (\log(x + 10) + x^{2})^{\epsilon}$,
(B) $y = \log(x + 10) + x^{2 \vert \epsilon\vert}$
(C) $y = \log(x^{7} + 5) + x^{5} - \sin(x^{2}\vert\epsilon\vert)$. As can be seen from Table (\ref{complex_acc}), LiNGAM and ANM do not perform very well which is to be expected as they rely on the assumption of additive noise. PNL, which assumes a invertible interaction between the covariates and noise, performs at or above chance level in almost all cases with very good performance under periodic noise. The non-parametric approach of IGCI has very good performance across all the functional and noise settings, while KCDC achieves perfect performance in all cases except under Gaussian and uniform noise for (A).

\begin{table}[!htbp]
\caption{Complex Noise: Classification Accuracies Over $100$ Datasets.}
\label{complex_acc}
\begin{center}
\begin{tabular}{ l r r r}
\hline
(A)  & Gaussian  & Uniform  & Exponential \\
\hline
LiNGAM    &  0\% & 2\% & 0\%  \\
ANM       &  28\% & 26\% & 24\%  \\
PNL       &  55\% & 50\% & 48\% \\
IGCI      &  {\bf 100\%} & 85\% & {\bf 100\%} \\
KCDC      &  98\% & {\bf 92\%} & {\bf 100\%} \\
\hline
(B)  & Gaussian  & Uniform  & Exponential \\
\hline
LiNGAM  &  31\% & 32\% & 23\% \\
ANM     &  16\% & 54\% & 6\%  \\
PNL     &  56\% & 50\% & 72\% \\
IGCI    &  88\% & 72\% & 97\%  \\
KCDC    & {\bf 100\%} & {\bf 100\%} & {\bf 100\%}  \\
\hline
(C)  & Gaussian  & Uniform  & Exponential \\
\hline
LiNGAM  & 0\% & 0\% & 1\%  \\
ANM     & 31\% & 19\% & 37\% \\
PNL     & 95\% & 92\% & 92\% \\
IGCI    & 97\% & 98\% & 98\%  \\
KCDC    & {\bf 100\%} & {\bf 100\%} & {\bf 100\%}  \\
\hline
\end{tabular}
\end{center}
\end{table}

\subsection{T\"{U}BINGEN CAUSE-EFFECT PAIRS}
Next, we discuss the performance of our method on real-world data. For this purpose, we test KCDC on the only widely used benchmark dataset T\"{u}bingen Cause-Effect Pairs (TCEP) \citep{mooij2015pairs}. This dataset is comprised of  real-world cause-effect samples that are collected across very diverse subject areas with the true causal direction provided by human experts. Due to the heterogenous origins of the data pairs, many diverse functional dependencies are expected to be present in TCEP.

In order to show the flexibility and capacity of KCDC when dealing with many diverse functional dependencies simultaneously, we test it using both the direct comparison decision rule and the majority decision rule. We use TCEP version 1.0 which consists of 100 cause-effect pairs. Each pair is assigned a weight in order to account for potential sources of bias given that different pairs are sometimes selected from the same multivariable dataset. Following the wide-spread approach present in the literature of testing only on scalar-valued pairs, we remove the multivariate pairs 52, 53, 54, 55 and 71 from TCEP in order to ensure a fair comparison to previous work. Note that contrary to some methods in literature, this is not necessary for our approach. For the majority approach, we choose the best settings of the kernel hyperparameters as inferred from the synthetic experiments. The direct approach represents the single best performing hyperparameter configuration on TCEP.

From the summary of classification accuracies of KCDC and related methods in Table \ref{res}, we see that KCDC is competitive to the state-of-the-art methods even when only one setting of kernel hyperparameters is used, i.e. when the direct comparison decision rule is used. When we combine multiple kernel hyperparameters under the majority vote approach, we see that our method outperforms other methods by a significant margin. Note that the review \citep{mooij2016distinguishing} discusses additive noise models \citep{hoyer2009nonlinear} and information-geometric causal inference \citep{daniusis2012inferring}. In particular, an extensive experimental evaluation of these methods across a wide range of hyperparameter settings is performed. In the fourth row of Table \ref{res}, we report the most favourable outcome across both types of methods of their large-scale experimental analysis. For testing RCC on TCEP v1.0, we use the code provided in \citep{lopez2015towards}.

\begin{table}[!htbp]
\centering
\caption{Classification Accuracy On TCEP}
\begin{tabular}{ l r }
 \hline
  Method & TCEP \\
 \hline
 ANM                                            & 59.5\%\\
 PNL                                            & 66.2\% \\
 RCC                                            & 64.67\% \\
 Best from \citep{mooij2016distinguishing}       & $\approx$ 74\% \\
 CGNN \citep{goudet2017causal}                   & 74.4\% \\
 KCDC (direct)                                  & 72.87\% \\
 KCDC (majority)                              & {\bf 78.71\%} \\
   \hline
\end{tabular}
\vspace{0.2cm}
\label{res}
\end{table}
\vspace{-0.1cm}

\subsection{INFERRING THE ARROW OF TIME}
In addition to the many real-world pairs above, we also test our method at inferring the direction of time on causal time series. Given a time series $\{X_{i}\}_{i=1}^{N}$, the task is to infer if $X_{i} \rightarrow X_{i+1}$ or $X_{i} \leftarrow X_{i+1}$.

We use a dataset containing quarterly growth rates of the real gross domestic product (GDP) of the UK, Canada and USA from 1980 to 2011 as in \citep{bauer2016arrow}. The resulting multivariate time series has length 124 and dimension three. According to the above selection of hyperparameters on the synthetic datasets, we chose a wide range of hyperparameters to test KCDC on. In particular, both on the response and input we used either a log kernel $k(x, x^{\prime}) = -\log(\norm{x - x^{\prime}}^{q} + 1)$ with $q$ in $[2,3,4]$ or an RBF kernel with bandwidth $[1, 1.5, 2]$ times the median heuristic. Across all of these hyperparameters, KCDC correctly identifies the causal direction with the confidence measure $\mathcal{T}^{KCDC}$ measuring the absolute relative difference between the KCDC measures varying between $2.45$ and $44565.6$. We compare our approach to methods readily applicable to causal infenrence on multivariable time series. In particular, LiNGAM does not identify the correct direction. On the other hand, the method developed in \citep{bauer2016arrow} that models the data as an autoregressive moving average model with non-Gaussian noise correctly identifies the causal direction.

\section{EXTENSIONS TO THE MULTIVARIABLE CASE}
While we present and discuss our method for the case of pairs of variables, it can be extended to the setting of more than two variables. Assuming we have $d$ variables with $d \geq 2$, i.e. $\mathbb{X} = \{X_{1}, \dots, X_{d}\}$, we can apply KCDC to every pair of variables $\{X_{i}, X_{j}\}\subseteq\mathbb{X}$ with $i\neq j$ while conditioning on all of the remaining variables in $\mathbb{X} \setminus \{X_{i}, X_{j}\}$. This corresponds to inferring the causal relationship between $X_{i}$ and $X_{j}$ while accounting for the confounding effect of all the remaining variables.

Another way of dealing with the multivariable setting is to use KCDC in conjunction with, for example, the PC algorithm \citep{spirtes2000causation}. In particular, one would first apply the PC algorithm to the data. The resulting DAG skeleton containing potentially many unoriented edges can then be processed with KCDC. In particular, our method can be applied sequentially to every pair of variables that is connected with an unoriented edge while conditioning on the remaining variables in the DAG.

Yet another approach to the multivariable case is to use KCDC measures as features in a multiclass classification problem for $d$-dimensional distributions.  However, as noted in \citep{lopez2015towards}, this approach quickly becomes rather cumbersome as the number of labels grows super-exponentially in the number of variables due to the rapid increase of the number of DAGs that can be constructed from $d$ variables.

\section{DISCUSSION}
In this paper, we proposed a fully nonparametric causal inference method that uses purely observational data and does not postulate \emph{a priori} assumptions on the functional relationship between the variables or the noise structure. As part of this, we developed a novel interpretation of the notion of asymmetry between cause and effect using information-theoretic considerations. In particular, we proposed to reason about this asymmetry in terms of the variability, across different values of the input, of the minimal description length of programs implementing the data-generating process of conditional distributions. As computing the Kolmogorov complexity is not tractable, we used the RKHS framework to construct highly expressive approximations in terms of the norm of conditional distribution embeddings. In order to quantify the description length variability, we proposed a flexible measure in terms of the within-set deviance of the RKHS norms of conditional mean embeddings. Based on this measure, we presented three decision rules for causal inference based on direct comparison, ensembling and classification, respectively. We extensively tested our proposed method across a wide range of diverse synthetic datasets showcasing its wide applicability. Furthermore, we tested our method on real-world time series data and the real-world benchmark dataset T\"{u}bingen Cause-Effect Pairs where we outperformed existing state-of-the-art methods by a significant margin.

Although we focused on conditional mean embeddings, there exist other representations of conditional distributions in the RKHS, e.g. conditional embedding operators.  The study of these representations and their comparison to KCDC is left for future work. As KCDC was developed under the assumption of causal sufficiency, extending it to explicitely model confounding is another possible avenue for future research.

\bibliographystyle{plainnat}
\bibliography{causality}

\end{document}